\documentclass[11pt,a4paper]{article}
\usepackage[hyperref]{acl2019}
\usepackage{times}
\usepackage{latexsym}

\usepackage{url}
\usepackage{units}
\usepackage{amstext}
\usepackage{amsmath}
\usepackage{amsfonts}
\usepackage{amssymb}
\usepackage{adjustbox}
\usepackage{booktabs}
\usepackage{todonotes}
\usepackage{multirow}
\usepackage{multicol}
\usepackage{hhline}
\usepackage{siunitx}
\usepackage{arydshln}
\usepackage{enumitem}
\setlist[itemize]{noitemsep}

\newcommand{\specialcell}[2][c]{%
  \begin{tabular}[#1]{@{}c@{}}#2\end{tabular}}

\aclfinalcopy 
\def\aclpaperid{655}

\title{A Wind of Change:\\ Detecting and Evaluating Lexical Semantic Change\\ across Times and Domains}

\author{Dominik Schlechtweg\textsuperscript{1}, Anna H\"atty\textsuperscript{1,2}, Marco del Tredici\textsuperscript{3}, Sabine Schulte im Walde\textsuperscript{1}\\
    \textsuperscript{1}Institute for Natural Language Processing, University of Stuttgart \\
	\textsuperscript{2}Robert Bosch GmbH, Corporate Research \\  
	\textsuperscript{3}Institute for Logic, Language and Computation, University of Amsterdam \\
	{\tt \{schlecdk,schulte\}@ims.uni-stuttgart.de}, \\
	{\tt anna.haetty@de.bosch.com, m.deltredici@uva.nl}}

\date{}

\begin{document}
\maketitle
\begin{abstract}
We perform an interdisciplinary large-scale evaluation for detecting lexical semantic divergences in a diachronic and in a synchronic task: semantic sense changes across time, and semantic sense changes across domains. Our work addresses the superficialness and lack of comparison in assessing models of diachronic lexical change, by bringing together and extending benchmark models on a common state-of-the-art evaluation task. In addition, we demonstrate that the same evaluation task and modelling approaches can successfully be utilised for the synchronic detection of domain-specific sense divergences in the field of term extraction.
\end{abstract}

\section{Introduction}
\label{sec:intro}

Diachronic \textit{Lexical Semantic Change (LSC)} detection, i.e., the automatic detection of word sense changes over time, is a flourishing new field within NLP \citep[i.a.]{Frermann:2016,Hamilton:2016,schlechtweg-EtAl:2017:CoNLL}.\footnote{An example for diachronic LSC is the German noun \textit{Vorwort} \citep{Paul02XXI}, which was mainly used in the meaning of `preposition' before $\approx$1800. Then \textit{Vorwort} rapidly acquired a new meaning `preface', which after 1850 has nearly exclusively been used.} Yet, it is hard to compare the performances of the various models, and optimal parameter choices remain unclear, because up to now most models have been compared on different evaluation tasks and data. Presently, we do not know which model performs best under which conditions, and if more complex model architectures gain performance benefits over simpler models. This situation hinders advances in the field and favors unfelicitous drawings of statistical laws of diachronic LSC \citep{dubossarsky2017}. 

In this study, we provide the first large-scale evaluation of an extensive number of approaches. Relying on an existing German LSC dataset we compare models regarding different combinations of semantic representations, alignment techniques and detection measures, while exploring various pre-processing and parameter settings. Furthermore, we introduce \textit{Word Injection} to LSC, a modeling idea drawn from term extraction, that overcomes the problem of vector space alignment. Our comparison of state-of-the-art approaches identifies best models and optimal parameter settings, and it suggests modifications to existing models which consistently show superior performance.

Meanwhile, the detection of lexical sense divergences across time-specific corpora is not the only possible application of LSC detection models. In more general terms, they have the potential to detect sense divergences between corpora of any type, not necessarily time-specific ones. We acknowledge this observation and further explore a \textit{synchronic LSC detection} task: identifying domain-specific changes of word senses in comparison to general-language usage, which is addressed, e.g., in term identification and automatic term extraction \cite{drouin2004,perez2016measuring,Haetty18LaymenStudy}, and in determining social and dialectal language variations \cite{del2017semantic,hovy/purschke:emnlp18}.\footnote{An example for domain-specific synchronic LSC is the German noun \textit{Form}. In general-language use, \textit{Form} means `shape'/`form', while in the cooking domain the predominant meaning is the domain-specific `baking tin'.}

For addressing the synchronic LSC task, we present a recent sense-specific term dataset \cite{haettySurel:2019} that we created analogously to the existing diachronic dataset, and we show that the diachronic models can be successfully applied to the synchronic task as well. This two-fold evaluation assures robustness and reproducibility of our model comparisons under various conditions.

\section{Related Work}
\label{sec:previous}

\paragraph{Diachronic LSC Detection.} Existing approaches for diachronic LSC detection are mainly based on three types of meaning representations: (i) semantic vector spaces, (ii) topic distributions, and (iii) sense clusters. In (i), semantic vector spaces, each word is represented as two vectors reflecting its co-occurrence statistics at different periods of time \citep[][]{Gulordava11,Kim14,Xu15,Eger:2016,HamiltonShiftDrift,Hamilton:2016,Hellrich16p2785,RosenfeldE18}. LSC is typically measured by the cosine distance (or some alternative metric) between the two vectors, or by differences in contextual dispersion between the two vectors \citep{kisselewetal:2016,schlechtweg-EtAl:2017:CoNLL}.
(ii) Diachronic topic models infer a probability distribution for each word over different word senses (or topics), which are in turn modeled as a distribution over words \citep[][]{Wang06,Bamman11p1,Wijaya11p35,Lau12p591,Mihalcea12,Cook14p1624,Frermann:2016}. LSC of a word is measured by calculating a novelty score for its senses based on their frequency of use.
(iii) Clustering models assign all uses of a word into sense clusters based on some contextual property \citep{Mitra15p773}. Word sense clustering models are similar to topic models in that they map uses to senses. Accordingly, LSC of a word is measured similarly as in (ii). For an overview on diachronic LSC detection, see \citet{2018arXiv181106278T}.

\paragraph{Synchronic LSC Detection.} We use the term synchronic LSC to refer to NLP research areas with a focus on how the meanings of words vary across domains or communities of speakers.
Synchronic LSC per se is not widely researched; for meaning shifts across domains, there is strongly related research which is concerned with domain-specific word sense disambiguation \cite{maynard1998term, chen2006context, Taghipour15, daille2016ambiguity} or term ambiguity detection \cite{Baldwin13TermAmbiguity,wang2013automatic}. The only notable work for explicitly measuring across domain meaning shifts is \citet{ferrari2017detecting}, which is based on semantic vector spaces and cosine distance.
Synchronic LSC across communities has been investigated as meaning variation in online communities, leveraging the large-scale data which has become available thanks to online social platforms \cite{del2017semantic,rotabi2017competition}.

\paragraph{Evaluation.} Existing evaluation procedures for LSC detection can be distinguished into evaluation on (i) empirically observed data, and (ii) synthetic data or related tasks. (i) includes case studies of individual words \citep[][]{Sagi09p104,Jatowt:2014,HamiltonShiftDrift}, stand-alone comparison of a few hand-selected words \citep[][]{Wijaya11p35,Hamilton:2016,del2017semantic}, comparison of hand-selected changing vs. semantically stable words \citep[][]{Lau12p591,Cook14p1624}, and post-hoc evaluation of the predictions of the presented models \citep[][]{Cook10,Kulkarni14,del2016tracing,Eger:2016,ferrari2017detecting}.
\citet{schlechtweg-EtAl:2017:CoNLL} propose a small-scale annotation of diachronic metaphoric change.

Synthetic evaluation procedures (ii) include studies that simulate LSC \citep{Cook10,Kulkarni14,RosenfeldE18}, evaluate sense assignments in WordNet \citep{Mitra15p773,Frermann:2016}, identify text creation dates, \citep[][]{Mihalcea12,Frermann:2016}, or predict the log-likelihood of textual data \citep[][]{Frermann:2016}.

Overall, the various studies use different evaluation tasks and data, with little overlap. Most evaluation data has not been annotated. Models were rarely compared to previously suggested ones, especially if the models differed in meaning representations. Moreover, for the diachronic task, synthetic datasets are used which do not reflect actual diachronic changes.

\section{Task and Data}
\label{sec:task}

Our study makes use of the evaluation framework proposed in \citet{Schlechtwegetal18}, where diachronic LSC detection is defined as a comparison between word uses in two time-specific corpora. We further applied the framework to create an analogous synchronic LSC dataset that compares word uses across general-language and domain-specific corpora. The common, meta-level task in our diachronic+synchronic setup is, given two corpora $C_a$ and $C_b$, to rank the targets in the respective datasets according to their degree of relatedness between word uses in $C_a$ and $C_b$.

\subsection{Corpora}
\label{subsec:corpora}

\paragraph{\textsc{DTA}} \citep{dta2017} is a freely available lemmatized, POS-tagged and spelling-normalized diachronic corpus of German containing texts from the 16th to the 20th century.

\vspace{-1mm}
\paragraph{\textsc{Cook}} is a domain-specific corpus. We crawled cooking-related texts from several categories (recipes, ingredients, cookware and cooking techniques) from the German cooking recipes websites \textit{kochwiki.de} and \textit{Wikibooks Kochbuch}\footnote{\url{de.wikibooks.org/wiki/Kochbuch}}.

\vspace{-1mm}
\paragraph{\textsc{SdeWaC}} \citep{faasseckart2012} is a cleaned version of the web-crawled corpus \textsc{deWaC} \citep{baroni2009wacky}. We reduced \textsc{SdeWaC} to \nicefrac{1}{8}th of its original size by selecting every 8th sentence for our general-language corpus.
\vspace{+2mm}\\
Table \ref{tab:corpSizes} summarizes the corpus sizes after applying pre-processing.
See Appendix \ref{sec:parameter} for pre-processing details.

\begin{table}[]
\centering
\begin{adjustbox}{width=0.45\textwidth}
\small
\begin{tabular}{lrrrr}
& \multicolumn{2}{c}{\textbf{Times}}&\multicolumn{2}{c}{\textbf{Domains}} \\
\hline
        & \multicolumn{1}{c}{\textsc{Dta18}} & \multicolumn{1}{c}{\textsc{Dta19}} 
        & \multicolumn{1}{c}{\textsc{SdeWaC}} & \multicolumn{1}{c}{\textsc{Cook}} \\
\hline
\textsc{L\textsubscript{all}}  & 26M  &  40M & 109M   & 1M     \\
\textsc{L/P\textsubscript{}}  & 10M  & 16M & 47M & 0.6M     \\
\end{tabular}
\end{adjustbox}
\caption{Corpora and their approximate sizes.}\label{tab:corpSizes}
\end{table}

\subsection{Datasets\footnote{The datasets are available in Appendix \ref{sec:supplemental} and at \url{https://github.com/Garrafao/LSCDetection}.} and Evaluation}
\label{subsec:datasets}

\paragraph{Diachronic Usage Relatedness (DURel).} DURel is a gold standard for diachronic LSC consisting of 22 target words with varying degrees of LSC \citep{Schlechtwegetal18}. Target words were chosen from a list of attested changes in a diachronic semantic dictionary \citep{Paul02XXI}, and for each target a random sample of use pairs from the DTA corpus was annotated for meaning relatedness of the uses on a scale from 1 (unrelated meanings) to 4 (identical meanings), both within and across the time periods 1750--1799 and 1850--1899. The annotation resulted in an average Spearman's $\rho=0.66$ across five annotators and 1,320 use pairs. For our evaluation of diachronic meaning change we rely on the ranking of the target words according to their mean usage relatedness across the two time periods. 

\paragraph{Synchronic Usage Relatedness (SURel).} SURel is a recent gold standard for synchronic LSC \citep{haettySurel:2019} using the same framework as in DURel. The 22 target words were chosen such as to exhibit different degrees of domain-specific meaning shifts, and use pairs were randomly selected from \textsc{SdeWaC} as general-language corpus and from \textsc{Cook} as domain-specific corpus. The annotation for usage relatedness across the corpora resulted in an average Spearman's $\rho=0.88$ across four annotators and 1,320 use pairs. For our evaluation of synchronic meaning change we rely on the ranking of the target words according to their mean usage relatedness between general-language and domain-specific uses.

\paragraph{Evaluation.} The gold LSC ranks in the DURel and SURel datasets are used to assess the correctness of model predictions by applying Spearman's rank-order correlation coefficient $\rho$ as evaluation metric, as done in similar previous studies \citep{Gulordava11,schlechtweg-EtAl:2017:CoNLL,SchlechtwegWalde18}. As corpus data underlying the experiments we rely on the corpora from which the annotated use pairs were sampled: \textsc{DTA} documents from 1750--1799 as $C_a$ and documents from 1850--1899 as $C_b$ for the diachronic experiments, and the \textsc{SdeWaC} corpus as $C_a$ and the \textsc{Cook} corpus as $C_b$ for the synchronic experiments.

\section{Meaning Representations\footnote{Find the hyperparameter settings in Appendix \ref{sec:parameter}. The scripts for vector space creation, alignment, measuring LSC and evaluation are available at \url{https://github.com/Garrafao/LSCDetection}.}}
\label{sec:repres}

Our models are based on two families of distributional meaning representations:  semantic vector spaces (Section~\ref{subsec:spaces}), and topic distributions (Section~\ref{subsec:topic}). All representations are bag-of-words-based, i.e. each word representation reflects a weighted bag of context words. The contexts of a target word $w_i$ are the words surrounding it in an $n$-sized window: $w_{i-n}, ..., w_{i-1}, w_{i+1}, ..., w_{i+n}$.

\subsection{Semantic Vector Spaces}
\label{subsec:spaces}

A semantic vector space constructed from a corpus $C$ with vocabulary $V$ is a matrix $M$, where each row vector represents a word $w$ in the vocabulary $V$ reflecting its co-occurrence statistics \citep{turney2010frequency}. We compare two state-of-the-art approaches to learn these vectors from co-occurrence data, (i) counting and (ii) predicting, and construct vector spaces for each time period and domain.

\subsubsection{Count-based Vector Spaces}

In a count-based semantic vector space the matrix $M$ is high-dimensional and sparse. The value of each matrix cell $M_{i,j}$ represents the number of co-occurrences of the word $w_i$ and the context $c_j$, $\#(w_i,c_j)$. In line with \citet{Hamilton:2016} we apply a number of transformations to these raw co-occurrence matrices, as previous work has shown that this improves results on different tasks \cite{Bullinaria2012,Levy2015}.

\vspace{+1mm}
\paragraph{Positive Pointwise Mutual Information (PPMI).} In PPMI representations the co-occurrence counts in each matrix cell $M_{i,j}$ are weighted by the positive mutual information of target $w_i$ and context $c_j$ reflecting their degree of association. The values of the transformed matrix are

\vspace{-3mm}
\small
\begin{equation*}
{ M}^{\textrm{PPMI}}_{i,j} = \max\left\lbrace\log\left(\frac{\#(w_i,c_j)\sum_c \#(c)^{\alpha}}{ \#(w_i)\#(c_j)^{\alpha}}\right)-\log(k),0\right\rbrace,
\end{equation*}
\normalsize
where $k > 1$ is a prior on the probability of observing an actual occurrence of $(w_i, c_j)$ and $0 < \alpha < 1$ is a smoothing parameter reducing PPMI's bias towards rare words \citep{Levy:2014,Levy2015}.

\vspace{+1mm}
\paragraph{Singular Value Decomposition (SVD).} Truncated SVD finds the optimal rank $d$ factorization of matrix $M$ with respect to L2 loss \citep{Eckart1936}. We use truncated SVD to obtain low-dimensional approximations of the PPMI representations by factorizing ${ M}^{\textrm{PPMI}}$ into the product of the three matrices ${ U}{ \Sigma}{ V}^\top$. We keep only the top $d$ elements of $\Sigma$ and obtain

\vspace{-2mm}
\begin{equation*}
{M}^{\textrm{SVD}} = { U_d}{ \Sigma^{p}_{d}},
\end{equation*}
where $p$ is an eigenvalue weighting parameter \cite{Levy2015}. The $i$th row of ${ M}^{\textrm{SVD}}$ corresponds to $w_i$'s $d$-dimensional representation. 

\vspace{+1mm}
\paragraph{Random Indexing (RI).} RI is a dimensionality reduction technique based on the Johnson-Lindenstrauss lemma according to which points in a vector space can be mapped into a randomly selected subspace under approximate preservation of the distances between points, if the subspace has a sufficiently high dimensionality \citep{Johnson1984,Sahlgren2004}. We reduce the dimensionality of a count-based matrix $M$ by multiplying it with a random matrix $R$:

\vspace{-2mm}
\begin{equation*}
{M}^{\textrm{RI}} = {M}{ R^{|\mathcal{V}| \times d}},
\end{equation*}
where the $i$th row of ${M}^{\textrm{RI}}$ corresponds to $w_i$'s $d$-dimensional semantic representation. The choice of the random vectors corresponding to the rows in $R$ is important for RI. We follow previous work \citep{Basile2015} and use sparse ternary random vectors with a small number $s$ of randomly distributed $-1$s and $+1$s, all other elements set to 0, and we apply subsampling with a threshold $t$.

\vspace{+1mm}
\subsubsection{Predictive Vector Spaces}

\vspace{+1mm}
\paragraph{Skip-Gram with Negative Sampling (SGNS)} differs from count-based techniques in that it directly represents each word $w \in V$ and each context $c \in V$ as a $d$-dimensional vector by implicitly factorizing $M=WC^\top$ when solving

\vspace{-3mm}
\small
\begin{equation*}
\arg\max_\theta \sum_{(w,c)\in D} \log \sigma(v_c \cdot v_w) + \sum_{(w,c) \in D'} \log \sigma (-v_c \cdot v_w),
\end{equation*}
\normalsize
where $\sigma(x) = \frac{1}{1+e^{-x}}$, $D$ is the set of all observed word-context pairs and $D'$ is the set of randomly generated negative samples \citep{Mikolov13a,Mikolov13b,GoldbergL14}. The optimized parameters $\theta$ are $v_{c_i}=C_{i*}$ and $v_{w_i}=W_{i*}$ for $w,c\in V$, $i\in 1,...,d$. $D'$ is obtained by drawing $k$ contexts from the empirical unigram distribution $P(c) = \frac{\#(c)}{|D|}$ for each observation of (w,c), cf. \citet{Levy2015}. SGNS and PPMI representations are highly related in that the cells of the implicitly factorized matrix $M$ are PPMI values shifted by the constant $k$ \citep{Levy:2014}. Hence, SGNS and PPMI share the hyper-parameter $k$. The final SGNS matrix is given by

\vspace{-2mm}
\begin{equation*}
{M}^{\textrm{SGNS}} = W,
\end{equation*}
where the $i$th row of ${M}^{\textrm{SGNS}}$ corresponds to $w_i$'s $d$-dimensional semantic representation. As in RI we apply subsampling with a threshold $t$. SGNS with particular parameter configurations has shown to outperform transformed count-based techniques on a variety of tasks \cite{marcobaroni2014predict,Levy2015}.

\vspace{+1mm}
\subsubsection{Alignment}

\vspace{+1mm}
\paragraph{Column Intersection (CI).} In order to make the matrices $A$ and $B$ from time periods $a < b$ (or domains $a$ and $b$) comparable, they have to be aligned via a common coordinate axis. This is rather straightforward for count and PPMI representations, because their columns correspond to context words which often occur in both $A$ and $B$ \citep{Hamilton:2016}. In this case, the alignment for $A$ and $B$ is

\begin{equation*}
\begin{split}
A_{*j}^{\textrm{CI}} = A_{*j}\textrm{~~ for all } c_j \in V_{a} \cap V_{b},\\
B_{*j}^{\textrm{CI}} = B_{*j}\textrm{~~ for all } c_j \in V_{a} \cap V_{b},\
\end{split}
\end{equation*}
where $X_{*j}$ denotes the $j$th column of $X$.

\vspace{+1mm}
\paragraph{Shared Random Vectors (SRV).} RI offers an elegant way to align count-based vector spaces and reduce their dimensionality at the same time \citep{Basile2015}. Instead of multiplying count matrices $A$ and $B$ each by a separate random matrix $R_A$ and $R_B$ they may be multiplied both by the same random matrix $R$ representing them in the same low-dimensional random space. Hence, $A$ and $B$ are aligned by
\begin{equation*}
\begin{split}
A^{\textrm{SVR}} = A R,\\
B^{\textrm{SVR}} = B R.
\end{split}
\end{equation*}
We follow \citeauthor{Basile2015} and adopt a slight variation of this procedure: instead of multiplying both matrices by exactly the same random matrix (corresponding to an intersection of their columns) we first construct a shared random matrix and then multiply $A$ and $B$ by the respective sub-matrix.

\vspace{+1mm}
\paragraph{Orthogonal Procrustes (OP).} In the low-dimensional vector spaces produced by SVD, RI and SGNS the columns may represent different coordinate axes (orthogonal variants) and thus cannot directly be aligned to each other. Following \citet{Hamilton:2016} we apply OP analysis to solve this problem. We represent the dictionary as a binary matrix D, so that $D_{i,j} = 1$ if $w_i \in V_b$ (the $i$th word in the vocabulary at time $b$) corresponds to $w_j \in V_a$. The goal is then to find the optimal mapping matrix $W^{*}$ such that the sum of squared Euclidean distances between $B$'s mapping
$B_{i*}W$ and $A_{j*}$ for the dictionary entries $D_{i,j}$ is minimized:
\begin{equation*}
W^{*} = \arg\min_{W} \sum_i\sum_j D_{i,j}\| B_{i*}W-A_{j*} \|^{2}.
\end{equation*}
Following standard practice we length-normalize and mean-center $A$ and $B$ in a pre-processing step \citep{artetxe2017acl}, and constrain $W$ to be orthogonal, which preserves distances within each time period. Under this constraint, minimizing the squared Euclidean distance becomes equivalent to maximizing the dot product when finding the optimal rotational alignment \citep{Hamilton:2016,artetxe2017acl}.
The optimal solution for this problem is then given by $W^{*}=UV^{\top}$, where $B^{\top} DA=U\Sigma V^{\top}$ is the SVD of $B^{\top} DA$. Hence, $A$ and $B$ are aligned by
\begin{equation*}
\begin{split}
&A^{\textrm{OP}} = A,\\
&B^{\textrm{OP}} = B W^{*},
\end{split}
\end{equation*}
where $A$ and $B$ correspond to their preprocessed versions. We also experiment with two variants: $\textrm{OP}_{-}$ omits mean-centering \citep{Hamilton:2016}, which is potentially harmful as a better solution may be found after mean-centering. $\textrm{OP}_{+}$ corresponds to OP with additional pre- and post-processing steps and has been shown to improve performance in research on bilingual lexicon induction \citep{artetxe2018aaai,artetxe2018acl}. We apply all OP variants only to the low-dimensional matrices.

\vspace{+1mm}
\paragraph{Vector Initialization (VI).} In VI we first learn $A^{\textrm{VI}}$ using standard SGNS and then initialize the SGNS model for learning $B^{\textrm{VI}}$ on $A^{\textrm{VI}}$ \citep{Kim14}. The idea is that if a word is used in similar contexts in $a$ and $b$, its vector will be updated only slightly, while more different contexts lead to a stronger update.

\vspace{+1mm}
\paragraph{Word Injection (WI).} Finally, we use the word injection approach by \citet{ferrari2017detecting} where target words are substituted by a placeholder in one corpus before learning semantic representations, and a single matrix $M^{\textrm{WI}}$ is constructed for both corpora after mixing their sentences. The advantage of this approach is that all vector learning methods described above can be directly applied to the mixed corpus, and target vectors are constructed directly in the same space, so no post-hoc alignment is necessary.

\vspace{+1mm}
\subsection{Topic Distributions}
\label{subsec:topic}

\vspace{+1mm}
\paragraph{Sense ChANge (SCAN).} SCAN models LSC of word senses via smooth and gradual changes in associated topics \citep{Frermann:2016}. The semantic representation inferred for a target word $w$ and time period $t$ consists of a $K$-dimensional distribution over word senses $\phi^{t}$ and a $V$-dimensional distribution over the vocabulary $\psi^{t,k}$ for each word sense $k$, where $K$ is a predefined number of senses for target word $w$. SCAN places parametrized logistic normal priors on $\phi^{t}$ and $\psi^{t,k}$ in order to encourage a smooth change of parameters, where the extent of change is controlled through the precision parameter $K^{\phi}$, which is learned during training. 

Although $\psi^{t,k}$ may change over time for word sense $k$, senses are intended to remain thematically consistent as controlled by word precision parameter $K^{\psi}$. This allows comparison of the topic distribution across time periods. For each target word $w$ we infer a SCAN model for two time periods $a$ and $b$ and take $\phi^{a}_w$ and $\phi^{b}_w$ as the respective semantic representations.

\vspace{+1mm}
\section{LSC Detection Measures}
\label{sec:measures}

LSC detection measures predict a degree of LSC from two time-specific semantic representations of a word $w$. They either capture the contextual similarity (Section~\ref{sec:similarity}) or changes in the contextual dispersion (Section~\ref{subsec:dispersion}) of $w$'s representations.\footnote{Find an overview of which measure was applied to which representation type in Appendix \ref{sec:parameter}.}

\vspace{+1mm}
\subsection{Similarity Measures}
\label{sec:similarity}

\vspace{+2mm}
\paragraph{Cosine Distance (CD).} CD is based on cosine similarity which measures the cosine of the angle between two non-zero vectors $\vec{x},\vec{y}$ with equal magnitudes \cite{salton1986introduction}:
	\begin{equation*}
	cos(\vec{x},\vec{y}) = \frac{\vec{x} \cdot \vec{y}}{\sqrt{\vec{x} \cdot \vec{x}} \sqrt{\vec{y} \cdot \vec{y}}}.
	\end{equation*}
The cosine distance is then defined as
	\begin{equation*}
	CD(\vec{x},\vec{y})=1-cos(\vec{x},\vec{y}).
	\end{equation*}
CD's prediction for a degree of LSC of $w$ between time periods $a$ and $b$ is obtained by $CD(\vec{w}_a,\vec{w}_b)$.

\vspace{+1mm}
\paragraph{Local Neighborhood Distance (LND).} LND computes a second-order similarity for two non-zero vectors $\vec{x},\vec{y}$ \citep{HamiltonShiftDrift}. It measures the extent to which $\vec{x}$ and $\vec{y}$~'s distances to their shared nearest neighbors differ. First the cosine similarity of $\vec{x},\vec{y}$ with each vector in the union of the sets of their $k$ nearest neighbors $N_k(\vec{x})$ and $N_k(\vec{y})$ is computed and represented as a vector $s$ whose entries are given by
    \begin{equation*}
    \begin{split}
    s(j) = \text{cos}(\vec{x},\vec{z}_j) \quad \forall \vec{z}_j\in N_{k}(\vec{x}) \cup N_{k}(\vec{y}).
    \end{split}
    \end{equation*}
LND is then computed as cosine distance between the two vectors:
	\begin{equation*}
    LND(\vec{x},\vec{y}) = CD(\vec{s_x}, \vec{s_y}).
	\end{equation*}
LND does not require matrix alignment, because it measures the distances to the nearest neighbors in each space separately. It was claimed to capture changes in paradigmatic rather than syntagmatic relations between words \citep{HamiltonShiftDrift}.

\paragraph{Jensen-Shannon Distance (JSD).} JSD computes the distance between two probability distributions $\phi_x,\phi_y$ of words $w_x, w_y$ \citep{Lin1991,DonosoS17}. It is the symmetrized square root of the Kullback-Leibler divergence:

\small
\begin{equation*}
JSD(\phi_x||\phi_y)=\sqrt{\frac{D_{KL}(\phi_x||M)+D_{KL}(\phi_y||M)}{2}}\,,
\end{equation*}
\normalsize
where $M=(\phi_x+\phi_y)/2$. JSD is high if $\phi_x$ and $\phi_y$ assign different probabilities to the same events.

\begin{table*}[]
	\center
	\small
	\hspace*{-10pt}
	\begin{tabular}{ c | c  S  c  c  c  c | c }
		\hline
		\textbf{Dataset} & \textbf{Preproc} & \textbf{Win} & \specialcell{\textbf{Space}} & \specialcell{\textbf{Parameters}\\}
		& \specialcell{\textbf{Align}\\} & \specialcell{\textbf{Measure}}  & 
		\textbf{Spearman  m (h, l)} \\ \hline
		
		\multirow{5}{*}{\textbf{DURel}} & \multirow{1}{*}{\textsc{L\textsubscript{all}}} &10& SGNS & k=1,t=None & $\textrm{OP}$ & CD & \textbf{0.866} \tiny (0.914, 0.816) \\ 
		&\multirow{1}{*}{\textsc{L\textsubscript{all}}} &10& SGNS & k=5,t=None & $\textrm{OP}$ & CD & 0.857 \tiny  (0.891, 0.830) \\ 
		&\multirow{1}{*}{\textsc{L\textsubscript{all}}} &5& SGNS & k=5,t=0.001 & $\textrm{OP}$ & CD & 0.835 \tiny (0.872, 0.814) \\ 
		&\multirow{1}{*}{\textsc{L\textsubscript{all}}} &10& SGNS & k=5,t=0.001 & $\textrm{OP}$ & CD & 0.826 \tiny (0.863, 0.768)  \\
		&\multirow{1}{*}{\textsc{L/P}} &2& SGNS & k=5,t=None & $\textrm{OP}$ & CD & 0.825 \tiny  (0.826, 0.818) \\ 
		\hline
		\multirow{5}{*}{\textbf{SURel}} & \textsc{L/P} &2& SGNS & k=1,t=0.001 & $\textrm{OP}$ & CD & \textbf{0.851} \tiny  (0.851, 0.851)  \\ 
		& \textsc{L/P} &2& SGNS & k=5,t=None & $\textrm{OP}$ & CD & 0.850 \tiny  (0.850, 0.850) 
		  \\ 
		& \textsc{L/P} &2& SGNS & k=5,t=0.001 & $\textrm{OP}$ & CD & 0.834 \tiny  (0.838, 0.828)   
		  \\ 
		 & \textsc{L/P} &2& SGNS & k=5,t=0.001 & $\textrm{OP}_{-}$ & CD & 0.831 \tiny  (0.836, 0.817) 
		  \\ 
		  & \textsc{L/P} &2& SGNS & k=5,t=0.001 & $\textrm{OP}$ & CD & 0.829 \tiny  (0.832, 0.823) 
		  \\  \hline  
	\end{tabular}
	\caption{Best results of $\rho$ scores (Win=Window Size, Preproc=Preprocessing, Align=Alignment, k=negative sampling, t=subsampling, Spearman m(h,l): mean, highest and lowest results).}
	\label{tab:bestResults}
\end{table*}

\vspace{+2mm}
\subsection{Dispersion Measures}
\label{subsec:dispersion}

\vspace{+2mm}
\paragraph{Frequency Difference (FD).} The log-transformed relative frequency of a word $w$ for a corpus $C$ is defined by
	\begin{equation*}
    F(w,C)=\log \frac{|w\in C|}{|C|}
	\end{equation*}
FD of two words $x$ and $y$ in two corpora $X$ and $Y$ is then defined by the absolute difference in F:
	\begin{equation*}
    FD(x,X,y,Y)= |F(x,X)-F(y,Y)|
	\end{equation*}
FD's prediction for $w$'s degree of LSC between time periods $a$ and $b$ with corpora $C_a$ and $C_b$ is computed as $FD(w,C_a,w,C_b)$ (parallel below). 

\vspace{+1mm}
\paragraph{Type Difference (TD).} TD is similar to FD, but based on word vectors $\vec{w}$ for words $w$. The normalized log-transformed number of context types of a vector $\vec{w}$ in corpus $C$ is defined by
	\begin{equation*}
    T(\vec{w},C) = \log \frac{\sum_{i=1} 1 \quad \textrm{if $\vec{w}_i \neq 0$}}{|C_T|} ,
	\end{equation*}
where $|C_T|$ is the number of types in corpus $C$. The TD of two vectors $\vec{x}$ and $\vec{y}$ in two corpora $X$ and $Y$ is the absolute difference in T:
	\begin{equation*}
    TD(\vec{x},X,\vec{y},Y)= |T(\vec{x},X)-T(\vec{y},Y)|.
	\end{equation*}

\vspace{+1mm}
\paragraph{Entropy Difference (HD).} HD relies on vector entropy as suggested by \citet{Santus:2014}. The entropy of a non-zero word vector $\vec{w}$ is defined by
	\begin{equation*}
    VH(\vec{w}) = -\sum_{i=1} \frac{\vec{w}_i}{\sum_{j=1} \vec{w}_j} \; \log \frac{\vec{w}_i}{\sum_{j=1} \vec{w}_j}.
	\end{equation*}
VH is based on Shannon's entropy \citep{Shannon:1948}, which measures the unpredictability of $w$'s co-occurrences \citep{schlechtweg-EtAl:2017:CoNLL}. HD is defined as
	\begin{equation*}
    HD(\vec{x},\vec{y}) = |VH(\vec{x})-VH(\vec{y})|.
	\end{equation*}
We also experiment with differences in H between topic distributions $\phi^{a}_w,\phi^{b}_w$, which are computed in a similar fashion, and with normalizing VH by dividing it by $\log(VT(\vec{w}))$, its maximum value. 

\vspace{+2mm}
\section{Results and Discussions}
\label{sec:results}

First of all, we observe that nearly all model predictions have a strong positive correlation with the gold rank. Table \ref{tab:bestResults} presents the overall best results across models and parameters.\footnote{For models with randomness we computed the average results of five iterations.} With $\rho=0.87$ for diachronic LSC (DURel) and $\rho=0.85$ for synchronic LSC (SURel), the models reach comparable and unexpectedly high performances on the two distinct datasets. The overall best-performing model is Skip-Gram with orthogonal alignment and cosine distance (SGNS+OP+CD). The model is robust in that it performs best on both datasets and produces very similar, sometimes the same results across different iterations.

\vspace{+1mm}
\paragraph{Pre-processing and Parameters.} Regarding pre-processing, the results are less consistent: \textsc{L\textsubscript{all}} (all lemmas) dominates in the diachronic task, while \textsc{L/P} (lemma:pos of content words) dominates in the synchronic task. In addition, \textsc{L/P} pre-processing, which is already limited on content words, prefers shorter windows, while \textsc{L\textsubscript{all}} (pre-processing where the complete sentence structure is maintained) prefers longer windows. Regarding the preference of \textsc{L/P} for SURel, we blame noise in the \textsc{Cook} corpus, which contains a lot of recipes listing ingredients and quantities with numerals and abbreviations, to presumably contribute little information about context words. For instance, \textsc{Cook} contains $4.6\%$ numerals, while \textsc{DTA} only contains $1.2\%$ numerals.

Looking at the influence of subsampling, we find that it does not improve the mean performance for Skip-Gram (SGNS) (with $\rho=0.506$, without $\rho=0.517$), but clearly for Random Indexing (RI) (with $\rho=0.413$, without $\rho=0.285$). \citet{Levy2015} found that SGNS prefers numerous negative samples ($k>1$), which is confirmed here: mean $\rho$ with $k=1$ is $0.487$, and mean $\rho$ with $k=5$ is $0.535$.\footnote{For PPMI we observe the opposite preference, mean $\rho$ with $k=1$ is $0.549$ and mean $\rho$ with $k=5$ is $0.439$.} This finding is also indicated in Table \ref{tab:bestResults}, where $k=5$ dominates the 5 best results on both datasets; yet, $k=1$ provides the overall best result on both datasets.

\paragraph{Semantic Representations.} Table \ref{tab:spaces} shows the best and mean results for different semantic representations. SGNS is clearly the best vector space model, even though its mean performance does not outperform other representations as clearly as its best performance. Regarding count models, PPMI and SVD show the best results.

SCAN performs poorly, and its mean results indicate that it is rather unstable. This may be explained by the particular way in which SCAN constructs context windows: it ignores asymmetric windows, thus reducing the number of training instances considerably, in particular for large window sizes.

\vspace{+1mm}
\begin{table}[htp]
	\center
	\small
	\begin{adjustbox}{width=0.8\linewidth}
	\begin{tabular}{ c | l  c  c }
		\hline
	\textbf{Dataset}&\textbf{Representation} &\textbf{best}&\textbf{mean}\\		
		\hline
		\multirow{6}{*}{\textbf{DURel}} & raw count & 0.639 & 0.395\\
		&  PPMI & 0.670 & 0.489\\
		&  SVD & 0.728 & 0.498\\
		&  RI & 0.601  & 0.374 \\ 
		&  SGNS & \textbf{0.866} & \textbf{0.502}\\ 
		& SCAN & 0.327 & 0.156 \\
		\hline 
		\multirow{6}{*}{\textbf{SURel}} & raw count & 0.599 & 0.120\\
		&  PPMI & 0.791 & 0.500\\
		&  SVD & 0.639 & 0.300\\
		&  RI & 0.622 & 0.299 \\ 
		&  SGNS & \textbf{0.851} & \textbf{0.520} \\
		& SCAN & 0.082 & -0.244 \\
		\hline
	\end{tabular}
	\end{adjustbox}
	\caption{Best and mean $\rho$ scores across similarity measures (CD, LND, JSD) on semantic representations.}
	\label{tab:spaces}
\end{table}

\paragraph{Alignments.} The fact that our modification of \citet{Hamilton:2016} (SGNS+OP) performs best across datasets confirms our assumption that column-mean centering is an important preprocessing step in Orthogonal Procrustes analysis and should not be omitted.

Additionally, the mean performance in Table \ref{tab:CompOP} shows that OP is generally more robust than its variants. $\textrm{OP}_{+}$ has the best mean performance on DURel, but performs poorly on SURel. \citet{artetxe2018aaai} show that the additional pre- and post-processing steps of $\textrm{OP}_{+}$ can be harmful in certain conditions. We tested the influence of the different steps and identified the non-orthogonal whitening transformation as the  main reason for a performance drop of $\approx$20\%.

In order to see how important the alignment step is for the low-dimensional embeddings (SVD/RI/SGNS), we also tested the performance without alignment (`None' in Table \ref{tab:CompOP}). As expected, the mean performance drops considerably. However, it remains positive, which suggests that the spaces learned in the models are not random but rather slightly rotated variants.

Especially interesting is the comparison of Word Injection (WI) where one common vector space is learned against the OP-models where two separately learned vector spaces are aligned. Although WI avoids (post-hoc) alignment altogether, it is consistently outperformed by OP, which is shown in Table \ref{tab:CompOP} for low-dimensional embeddings.\footnote{We see the same tendency for WI against random indexing with a shared random space (SRV), but instead variable results for count and PPMI alignment (CI). This contradicts the findings in \citet{Dubossarskyetal19}, using, however, a different task and synthetic data.} We found that OP profits from mean-centering in the pre-processing step: applying mean-centering to WI matrices improves the performance by 3\% on WI+SGNS+CD.

The results for Vector Initialization (VI) are unexpectedly low (on DURel mean $\rho=-0.017$, on SURel mean $\rho=0.082$). An essential parameter choice for VI is the number of training epochs for the initialized model. We experimented with 20 epochs instead of 5, but could not improve the performance. This contradicts the results obtained by \citet{Hamilton:2016} who report a ``negligible'' impact of VI when compared to $\textrm{OP}_{-}$. We reckon that VI is strongly influenced by frequency. That is, the more frequent a word is in corpus C$_b$, the more its vector will be updated after initialization on C$_a$. Hence, VI predicts more change with higher frequency in C$_b$.

\vspace{+1mm}
\begin{table}[t]
	\centering
	\begin{adjustbox}{width=0.45\textwidth}
	\begin{tabular}{c|r r r |r |r }
	    \hline
\textbf{Dataset}&\textbf{$\textrm{OP}$}&\textbf{$\textrm{OP}_{-}$}&\textbf{$\textrm{OP}_{+}$}&\textbf{WI}&\textbf{None} \\
            \hline
            \textbf{DURel} &0.618&0.557&\textbf{0.621}&0.468&0.254\\ 
            \textbf{SURel} &\textbf{0.590}&0.514&0.401&0.492&0.285\\ 
		\hline
	\end{tabular}
	\end{adjustbox}
	\caption{Mean $\rho$ scores for CD across the alignments. Applies only to RI, SVD and SGNS.}
	\label{tab:CompOP}
\end{table}

\vspace{-1mm}
\paragraph{Detection Measures.} Cosine distance (CD) dominates Local Neighborhood Distance (LND) on all vector space and alignment types (e.g., mean $\rho$ on DURel with SGNS+OP is $0.723$ for CD vs. $0.620$ for LND) and hence should be generally preferred if alignment is possible. Otherwise LND or a variant of WI+CD should be used, as they show lower but robust results.\footnote{JSD was not included here, as it was only applied to SCAN and its performance thus strongly depends on the underlying meaning representation.} Dispersion measures in general exhibit a low performance, and previous positive results for them could not be reproduced \citep{schlechtweg-EtAl:2017:CoNLL}. It is striking that, contrary to our expectation, dispersion measures on SURel show a strong negative correlation (max. $\rho=-0.79$). We suggest that this is due to frequency particularities of the dataset: SURel's gold LSC rank has a rather strong negative correlation with the targets' frequency rank in the \textsc{Cook} corpus ($\rho=-0.51$). Moreover, because \textsc{Cook} is magnitudes smaller than \textsc{SdeWaC} the normalized values computed in most dispersion measures in \textsc{Cook} are much higher. This gives them also a much higher weight in the final calculation of the absolute differences. Hence, the negative correlation in \textsc{Cook} propagates to the final results. This is supported by the fact that the only measure not normalized by corpus size (HD) has a positive correlation. As these findings show, the dispersion measures are strongly influenced by frequency and very sensitive to different corpus sizes. 

\paragraph{Control Condition.} As we saw, dispersion measures are sensitive to frequency. Similar observations have been made for other LSC measures \citep{dubossarsky2017}. In order to test for this influence within our datasets we follow \citet{dubossarsky2017} in adding a control condition to the experiments for which sentences are randomly shuffled across corpora (time periods). For each target word we merge all sentences from the two corpora $C_a$ and $C_b$ containing it, shuffle them, split them again into two sets while holding their frequencies from the original corpora approximately stable and merge them again with the original corpora. This reduces the target words' mean degree of LSC between $C_a$ and $C_b$ significantly. Accordingly, the mean degree of LSC predicted by the models should reduce significantly if the models measure LSC (and not some other controlled property of the dataset such as frequency). We find that the mean prediction on a result sample (\textsc{L/P}, win=2) indeed reduces from $0.5$ to $0.36$ on DURel and from $0.53$ to $0.44$ on SURel. Moreover, shuffling should reduce the correlation of individual model predictions with the gold rank, as many items in the gold rank have a high degree of LSC, supposedly being canceled out by the shuffling and hence randomizing the ranking. Testing this on a result sample (SGNS+OP+CD, \textsc{L/P}, win=2, k=1, t=None), as shown in Table \ref{tab:shuffle}, we find that it holds for DURel with a drop from $\rho=0.816$ (ORG) to $0.180$ on the shuffled (SHF) corpora, but not for SURel where the correlation remains stable ($0.767$ vs. $0.763$). We hypothesize that the latter may be due to SURel's frequency properties and find that downsampling all target words to approximately the same frequency in both corpora ($\approx 50$) reduces the correlation (+DWN). However, there is still a rather high correlation left ($0.576$). Presumably, other factors play a role: (i) Time-shuffling may not totally randomize the rankings because words with a high change still end up having slightly different meaning distributions in the two corpora than words with no change at all. Combined with the fact that the SURel rank is less uniformly distributed than DURel this may lead to a rough preservation of the SURel rank after shuffling. (ii) For words with a strong change the shuffling creates two equally polysemous sets of word uses from two monosemous sets. The models may be sensitive to the different variances in these sets, and hence predict stronger change for more polysemous sets of uses.
Overall, our findings demonstrate that much more work has to be done to understand the effects of time-shuffling as well as sensitivity effects of LSC detection models to frequency and polysemy.

\begin{table}[t]
	\centering
	\begin{adjustbox}{width=0.65\linewidth}
	\begin{tabular}{c|r| rr }
	    \hline
             \textbf{Dataset} & \textbf{ORG}  & \textbf{SHF} & \textbf{+DWN}  \\ \hline
            \textbf{DURel} & \textbf{0.816} & 0.180 & 0.372  \\
            \textbf{SURel} & \textbf{0.767} & 0.763 & 0.576  \\ \hline
	\end{tabular}
	\end{adjustbox}
	\caption{$\rho$ for SGNS+OP+CD (\textsc{L/P}, win=2, k=1, t=None) before (ORG) and after time-shuffling (SHF) and downampling them to the same frequency (+DWN).}
	\label{tab:shuffle}
\end{table}

\vspace{+1mm}
\section{Conclusion}
\label{sec:conclusion}

We carried out the first systematic comparison of a wide range of LSC detection models on two datasets which were reliably annotated for sense divergences across corpora. The diachronic and synchronic evaluation tasks we introduced were solved with impressively high performance and robustness. We introduced \textit{Word Injection} to overcome the need of (post-hoc) alignment, but find that Orthogonal Procrustes yields a better performance across vector space types.

The overall best performing approach on both data suggests to learn vector representations for different time periods (or domains) with SGNS, to align them with an orthogonal mapping, and to measure change with cosine distance. We further improved the performance of the best approach with the application of mean-centering as an important pre-processing step for rotational vector space alignment.

\section*{Acknowledgments}

The first author was supported by the Konrad Adenauer Foundation and the CRETA center funded by the German Ministry for Education and Research (BMBF) during the conduct of this study. We thank Haim Dubossarsky, Simon Hengchen, Andres Karjus, Barbara McGillivray, Cennet Oguz, Sascha Schlechtweg, Nina Tahmasebi and the three anonymous reviewers for their valuable comments. We further thank Michael Dorna and Bingqing Wang for their helpful advice. We also thank Lea Frermann for providing the code of SCAN and helping to set up the implementation.

\bibliography{wind-of-change}
\bibliographystyle{acl_natbib}

\clearpage
\appendix

\section{Pre-processing and Hyperparameter Details}
\label{sec:parameter}

\paragraph{Corpora.} For our experiments we used the TCF-version of \textsc{DTA} released September 1, 2017.\footnote{\url{http://www.deutschestextarchiv.de/download}} For all corpora, we removed words below a frequency threshold $t$. For the smallest corpus \textsc{Cook} we set $t=2$, and set the other thresholds in the same proportion to the corpus size. This led to $t=25,37,97$ for \textsc{DTA18}, \textsc{DTA19} and \textsc{SdeWaC} respectively. (Note that we excluded three targets from the DURel dataset and one target from the SURel dataset because they were below the frequency threshold.) We then created two versions: 
\begin{itemize}
    \item a version with minimal pre-processing, i.e., with punctuation removed and lemmatization (\textsc{L\textsubscript{all}})
    \item a stronger preprocessed version with only content words. After punctuation removal, lemmatization and POS-tagging, only nouns, verbs and adjectives were retained in the form \textit{lemma:POS} (\textsc{L/P\textsubscript{}})
\end{itemize}

\paragraph{Context window.} For all models we experimented with values $n=\{2,5,10\}$ as done in \citet{Levy2015}. It is important to note that the extraction of context words differed between models, because of inherent parameter settings of the implementations. While our implementations of the count-based vectors have a stable window of size $n$, SGNS has a dynamic context window with maximal size $n$ \citep[cf.][]{Levy2015} and SCAN has as stable window of size $n$, but ignores all occurrences of a target word where the number of context words on either side is smaller than $n$. This may affect the comparability of the different models, as especially the mechanism of SCAN can lead to very sparse representations on corpora with small sentence sizes, as e.g. the \textsc{Cook} corpus. Hence, this variable should be controlled in future experiments.

\paragraph{Vector Spaces.} We followed previous work in setting further hyper-parameters \citep{Hamilton:2016,Levy2015}. We set the number of dimensions $d$ for SVD, RI and SGNS to 300. We trained all SGNS with 5  epochs. For PPMI we set $\alpha = .75$ and experimented with $k=\{1,5\}$ for PPMI and SGNS. For RI and SGNS we experimented with $t=\{none,.001\}$. For SVD we set $p=0$. In line with \citet{Basile2015} we set $s=2$ for RI and SRV. Note though that we had a lower $d$ than \citeauthor{Basile2015}, who set $d=500$.

\paragraph{SCAN.} We experimented with $K=\{4,8\}$. For further parameters we followed the settings chosen by \citet{Frermann:2016}: $K^\psi=10$ (a high value forcing senses to remain thematically consistent across time). We set $K^\phi$ = 4, and the Gamma parameters $a = 7$ and $b = 3$. We used $1,000$ iterations for the Gibbs sampler and set the minimum amount of contexts for a target word per time period $min=0$ and the maximum amount to $max=2000$.

\paragraph{Measures.} For LND we set $k=25$ as recommended by \citet{HamiltonShiftDrift}. The normalization constants for FD, HD and TD were calculated on the full corpus with the respective preprocessing (without deleting words below a frequency threshold).

\section{Model Overview}
\label{sec:modeloverview}

Find an overview of all tested combinations of semantic representations, alignments and measures in Table \ref{tab:reprealign}.

\begin{table*}[]
	\center
	\begin{adjustbox}{width=0.7\linewidth}
	\begin{tabular}{ l | c c c c c | c c c c c c}
		\hline
		 \multirow{2}{*}{\textbf{Semantic Representation}} & \multicolumn{5}{c|}{\textbf{Alignment}} & \multicolumn{6}{c}{\textbf{Measure}} \\
		 & CI & SRV & OP & VI & WI & CD & LND & JSD & FD & TD & HD\\
		\hline
		raw count & x &  &  &  & x & x & x &  &  & x & x \\
		PPMI & x &  &  &  & x & x & x &  &  &  & \\
                SVD &  &  & x &  & x & x & x &  &  &  & \\
		RI &  & x & x &  & x & x & x &  &  &  &  \\ 
		SGNS &  &  & x & x & x & x & x &  &  &  & \\ 
		SCAN &  &  &  &  &  &  &  & x &  &  & (x) \\
		\hline
	\end{tabular}
	\end{adjustbox}
	\caption{Combinations of semantic representation, alignment types and measures. (FD has been computed directly from the corpus.)}
	\label{tab:reprealign}
\end{table*}

\section{Datasets}
\label{sec:supplemental}

Find the datasets with the target words and their annotated degree of LSC in Tables \ref{tab:durel} and \ref{tab:surel}.

\begin{table}[]
	\center
	\small
	\begin{adjustbox}{width=0.95\linewidth}
	\begin{tabular}{ l l | r| r  r}
		\hline
		\multirow{2}{*}{\textbf{lexeme}}&\multirow{2}{*}{\textbf{POS}}&\multirow{2}{*}{\textbf{LSC}}&\multirow{2}{*}{\textbf{freq. C$_a$}}&\multirow{2}{*}{\textbf{freq. C$_b$}} \\
		 & & &  &   \\
		\hline
        Vorwort & NN & -1.58 & 85 & 273\\
        Donnerwetter & NN & -1.84 & 100 & 89\\
        Presse & NN & -1.88 & 193 & 1519\\
        Feine & NN & -1.93 & 112 & 84\\
        Anstalt & NN & -2.07 & 425 & 911\\
        Feder & NN & -2.14 & 1489 & 3022\\
        billig & ADJ & -2.43 & 2073 & 1705\\
        Motiv & NN & -2.66 & 104 & 2551\\
        Anstellung & NN & -2.68 & 53 & 499\\
        packen & VV & -2.74 & 279 & 1057\\
        locker & ADJ & -2.84 & 454 & 769\\
        technisch & ADJ & -2.89 & 25 & 2177\\
        geharnischt & ADJ & -3.0 & 56 & 117\\
        Zufall & NN & -3.11 & 2444 & 1618\\
        Bilanz & NN & -3.2 & 51 & 58\\
        englisch & ADJ & -3.34 & 1921 & 7280\\
        Reichstag & NN & -3.45 & 609 & 1781\\
        Museum & NN & -3.73 & 414 & 1827\\
        Abend & NN & -3.79 & 4144 & 4372\\
		\hline
	\end{tabular}
	\end{adjustbox}
	\caption{DURel dataset without \emph{flott}, \emph{Kinderstube} and \emph{Steckenpferd}, which were excluded for low frequency. C$_a$=\textsc{Dta18}, C$_b$=\textsc{Dta19}. LSC denotes the inverse compare rank from \citep{Schlechtwegetal18}, where high values mean high change.}	\label{tab:durel}
\end{table}

\begin{table}[]
	\center
	\small
	\hspace*{-10pt}
	\begin{adjustbox}{width=0.95\linewidth}
	\begin{tabular}{ l l | r|r r}
		\hline
		\multirow{2}{*}{\textbf{lexeme}}&\multirow{2}{*}{\textbf{POS}}&\multirow{2}{*}{\textbf{LSC}}&\multirow{2}{*}{\textbf{freq. C$_a$}}&\multirow{2}{*}{\textbf{freq. C$_b$}} \\
		 & & &  &   \\
		 \hline
        Schnee & NN & -1.05 & 2228 & 53\\
        Strudel & NN & -1.05 & 232 & 46\\
        schlagen & VV & -1.1 & 14693 & 309\\
        Gericht & NN & -1.15 & 13263 & 1071\\
        Schu{\ss} & NN & -1.42 & 2153 & 117\\
        Hamburger & NN & -1.53 & 5558 & 46\\
        abschrecken & VV & -1.75 & 730 & 170\\
        Form & NN & -2.25 & 36639 & 851\\
        trennen & VV & -2.65 & 5771 & 170\\
        Glas & NN & -2.7 & 3830 & 863\\
        Blech & NN & -2.95 & 409 & 145\\
        Prise & NN & -3.1 & 370 & 622\\
        Paprika & NN & -3.33 & 377 & 453\\
        Mandel & NN & -3.45 & 402 & 274\\
        Messer & NN & -3.5 & 1774 & 925\\
        Rum & NN & -3.55 & 244 & 181\\
        Salz & NN & -3.74 & 3087 & 5806\\
        Eiwei{\ss} & NN & -3.75 & 1075 & 3037\\
        Schokolade & NN & -3.98 & 947 & 251\\
        Gem\"use & NN & -4.0 & 2696 & 1224\\
        Schnittlauch & NN & -4.0 & 156 & 247\\
		\hline
	\end{tabular}
	\end{adjustbox}
	\vspace*{-7pt}
	\caption{SURel dataset without \emph{Messerspitze}, which was excluded for low frequency. C$_a$=\textsc{SdeWaC}, C$_b$=\textsc{Cook}. LSC denotes the inverse compare rank from \citep{Schlechtwegetal18}, where high values mean high change.}
	\label{tab:surel}
\end{table}

\end{document}